# Toward expert-level motivational interviewing for health behavior improvement with LLMs

Run-ze Hu[1], Yang Yang[1], Yi-hang Yang[1], Jing-qi Kong[1], Jia-hui Luo[2], Wen-yu Yang[3], Jing Chen[1], Jing-yao Liu[1], Hui-qun Zeng[1], Lei Zhang[4#], Zheng Liu[1#]

**Affiliations:**

1 Department of Maternal and Child Health, School of Public Health, Peking University, Beijing, China;

2 School of Public Health, Peking University, Beijing, China;

3 Peking University China Center For Health Development Studies, School of Public Health, Peking University, Beijing, China;

4 Taomi AI4Health Lab, Beijing, China.

**Corresponding to:**

Zheng Liu, Department of Maternal and Child Health, School of Public Health, Peking University, Beijing, China, liuzheng@bjmu.edu.cn, https://orcid.org/0000-0002-0405-2348.

Lei Zhang, Taomi AI4Health Lab, Beijing, China, 400.zhang@163.com.

**Data availability:**

Upon publication, all code and processed data necessary to reproduce the main findings will be made publicly available at https://github.com/CbyerDragon/MI-LLM.

# Abstract

**Background:** Motivational interviewing (MI) is an effective counseling approach for promoting health behavior change, but its impact is constrained by the need for highly trained human counselors.

**Objective:** This study aimed to explore a scalable alternative by developing and evaluating Large Language Models for Motivational Interviewing (MI-LLMs).

**Methods:** We first curated five Chinese psychological counseling corpora and, using GPT-4 with an MI-informed prompt, transcribed multi-turn dialogues from the two highest-quality datasets (CPsyCounD and PsyDTCorpus) into 2,040 MI-style counseling conversations, of which 2,000 were used for training and 40 for testing. Three Chinese-capable open-source LLMs (Baichuan2-7B-Chat, ChatGLM-4-9B-Chat and Llama-3-8B-Chinese-Chat-v2) were fine-tuned on this corpus and were named as MI-LLMs. We evaluated MI-LLMs using round-based automatic metrics and expert manual coding with the Motivational Interviewing Treatment Integrity (MITI) Coding Manual 4.2.1.

**Results:** Across all three models, fine-tuning substantially improved BLEU-4 and ROUGE scores compared with the base models, and manual coding showed that MI-LLMs achieved technical and relational global scores, and MI-adherent ratios that approached those of real MI dialogues, although complex reflections and reflection-to-question ratios remained less frequent.

**Conclusions:** These findings provide initial evidence that MI-oriented fine-tuning can endow general-purpose LLMs with core MI-consistent counseling behaviors, suggesting a scalable pathway toward AI-assisted health behavior change support while underscoring the need for further work on data scale, complex MI skills and real-world intervention trials.

Chronic diseases such as cardiovascular disease, diabetes, and obesity have become significant global health challenges[1, 2]. Health behavior promotion is fundamental to the prevention and management of most of these chronic diseases. However, achieving long-term maintenance of behavior change through such interventions remains challenging[3-8]. One line of research highlights the importance of enhancing individuals' intrinsic motivation[9].

To this end, motivational interviewing (MI), as an evidence-based counseling approach to facilitating behavior change[10], has been proven effective in promoting health behavior change across a variety of behaviors[11, 12], such as smoking cessation[13], weight loss[14], substance use[15], physical activity[16, 17], and chronic disease self-management[18]. However, MI's effectiveness depends strongly on the professional therapist's experiences and skills, which require extensive training and continuous practice over several years[19, 20]. These demanding training requirements not only constrain the scalability of MI, but also make it difficult for practitioners to maintain MI treatment integrity over time[21].

The rise of Large Language Models (LLMs) such as GPT-4 has introduced a promising tool for addressing these challenges. LLMs have demonstrated advanced natural language processing capabilities and have been successfully applied in a variety of health–related tasks[22, 23]. However, existing research on the integration of MI with artificial intelligence or conversational agents faces several key limitations, including small, self-constructed datasets[24], rule-based system architectures[25-27], and outcome evaluation not adherent to standardized MI fidelity frameworks[26, 28].

In light of these limitations, the present study aimed to (1) provide a scalable pathway for constructing a Chinese MI-style dialogue dataset, (2) train fine-tuned LLMs to master the core skills of MI, and (3) assess the MI abilities of fine-tuned LLMs using both automatic and manual evaluation based on the Motivational Interviewing Treatment Integrity (MITI) Coding Manual 4.2.1[29].

# Methods

## Study design

Figure 1 shows the overall design of this study. This study first collected available Chinese psychological counseling dialogue datasets through online searches, randomly sampled them, and conducted automated dialogue quality evaluations to screen out datasets with higher scores. Due to the lack of structured content of motivational interviewing in the selected high-quality Chinese multi-round psychological counseling dialogue datasets, this study designed a prompt based on motivational interviewing and used GPT-4 to transcribe ordinary psychological dialogues into conversations based on motivational interviewing, thereby constructing a multi-round psychological dialogue dataset based on motivational interviewing. Using this dataset, we fine-tuned the large language model and developed the MI-LLM model. Finally, through automatic evaluation and manual evaluation, we comprehensively evaluated whether the MI-LLMs initially had the potential to promote healthy behaviors by conducting motivational interviewing.

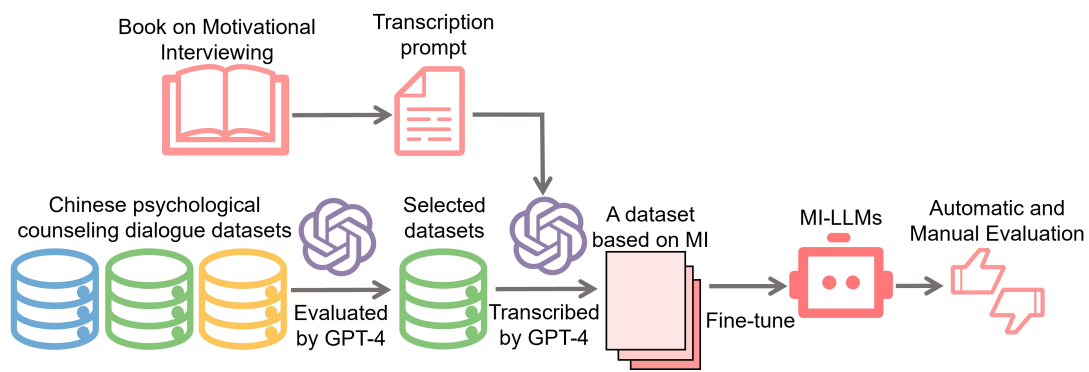

**Figure 1** Overall design of this study

## Screening and evaluating the counseling dialogue dataset

First, we searched Chinese psychological counseling dialogue datasets from platforms including Kaggle, GitHub, Hugging Face, and OpenDataLab. Then we directly downloaded or purchased them. After that, we randomly selected 50 conversations from each dataset and referenced CPsyCounD's automated evaluation methodology to evaluate the quality of these conversations[30]. Specifically, this evaluation criterion included four aspects: comprehensiveness, professionalism, authenticity, and safety. Details of these four aspects of evaluation criteria are shown in Supplemental Table 1. In the interest of data diversity, we decided to move forward with the top two datasets based on their aggregate scores.

**Transcription of the counseling dialogues**

According to an authoritative book in MI field[10], we summarized the four basic tasks and interview techniques of MI to prepare a transcription prompt for GPT-4 that could transcribe the Chinese psychological conversation into MI-style dialogues. The transcription prompt template was strictly developed based on prompt engineering principles[31], ensuring both structural rigor and functional appropriateness. Specifically, it was structured around the following key components: Role, Task Objective, the Four Core Tasks of Motivational Interviewing, Key Techniques of Motivational Interviewing, Transformation Steps, Guiding Principles, and Output Format Example. The specific prompt was shown in Supplemental Note 1 and Note 2.

**Original model information**

Three open-source LLMs with good Chinese language capability, Baichuan2-7B-Chat, ChatGLM-4-9B-Chat, and Llama-3-8B-Chinese-Chat-v2, were downloaded from Hugging face as models for fine-tuning and testing.

Baichuan2-7B-Chat[32] is an open-source large language model launched by Baihchuan Intelligence in December 2023, which was trained with a high-quality corpus of 2.6 trillion tokens.

ChatGLM-4-9B-Chat[33] is an open-source version of the pre-trained model in the GLM-4 series launched by Smart Spectrum AI in June 2024 and has demonstrated superior performance beyond Meta-Llama-3-8B in multi-faceted dataset measurements.

Meta-Llama-3-8B[34] is an open-source LLM released by Meta in April 2024, using a high-quality corpus of 15 trillion tokens for pre-training with excellent performance. However, since only 5% of the training corpus of Meta-Llama-3-8B is non-English, it often answers Chinese prompts in English during inference. Instead of using the original model of Llama3, we use Llama-3-8B-Chinese-Chat-v2[35], a model based on Meta-Llama-3-8B, which provides a further 100,000 tokens of Chinese corpus for preference training, so that the model's Chinese language ability has been greatly developed.

To evaluate the models' language comprehension and reasoning capabilities, we compiled models' scores on three key benchmarks as follows.

MMLU (Massive Multitask Language Understanding): A large-scale multi-task benchmark mainly used for evaluating English language models across 57 diverse tasks, measuring broad language understanding and reasoning abilities.

CMMLU (Chinese Multi-task Language Understanding): A multi-task benchmark for evaluating comprehensive Chinese language understanding, including commonsense reasoning, mathematical reasoning, and sentiment analysis.

C-Eval (Chinese Evaluation): A benchmark designed to assess Chinese language understanding and generation capabilities. It covers multiple tasks such as reading comprehension, reasoning, and translation.

All benchmarks consist solely of multiple-choice questions. Model performance is reported as accuracy, therefore higher scores indicate better model capability. For better comparison, we also collected the publicly reported scores of GPT-4 and Meta Llama-3-8B on these benchmarks.

The specific scores for each model are shown in Table 1. The results showed that ChatGLM-4-9B-Chat performs strongly on C-Eval and CMMLU, demonstrating superior capability in Chinese-language applications. Although Llama-3-8B-Chinese-Chat-v2 and Baichuan2-7B-Chat achieved relatively lower scores, they still exhibited sufficient Chinese comprehension and reasoning abilities to serve as base models for fine-tuning in this study.

**Table 1** Models' scores on three key benchmarks and corresponding testing method

| Model | MMLU | CMMLU | C-Eval | Evaluation setting | Source |
|---|---|---|---|---|---|
| Baichuan2-7B-Chat[a] | 52.9 | 55.0 | 54.4 | 5-shot | [32] |
| ChatGLM-4-9B-Chat[a] | 72.4 | 75.1 | 75.6 | Not mentioned | [33] |
| Llama-3-8B-Chinese-Instruct-v2[a] | 63.7 | 52.4 | 49.8 | 5-shot | [36] |
| GPT4[b] | 83.9 | 70.3 | 68.4 | 5-shot[c] | [32] |
| Meta Llama-3-8B[b] | 65 | 50.8 | 49.4 | 5-shot | [36] |

[a] Models fine-tuned in this study.

[b] Models included for comparison.

[c] 5-shot means that when the model performs a task, it is provided with five examples as references or learning guides. These examples help the model recognize patterns or rules in the task, thereby improving its reasoning and predictions.

**Methods of fine-tuning**

Our study used the integrated tool LLaMA-Factory[37] which was designed specifically for LLM. It was a training framework that significantly simplified the usage process by encapsulating and optimizing various tools. This framework could enable users to accomplish tasks such as fine-tuning, evaluating, and deploying large models more conveniently.

This study adopted Low-Rank Adaptation (LoRA) to fine-tune the LLMs. The core idea of LoRA was to achieve model performance improvement by injecting trainable layers on top of the original large model. LoRA achieves efficient fine-tuning using less memory consumption by approximating the incremental parameters obtained from full parameter fine-tuning of LLM with fewer training parameters. During training, we set the following key hyperparameters: a learning rate of 1.0e-4, which is gradually reduced using a cosine learning rate scheduler; a per-device training batch size of 1, and a gradient accumulation step size of 8 to achieve a larger effective batch size with limited memory. The number of training epochs was set to 3 to avoid overfitting; the warmup ratio was set to 0.1 to gradually increase the learning rate; and BF16 precision was used to accelerate training and reduce memory usage. In the training process, logs were recorded every 10 steps, the model was saved every 500 steps, and the change curve of the loss function was plotted.

Through fine-tuning the three models—Baichuan2-7B-Chat, ChatGLM-4-9B-Chat, and Llama-3-8B-Chinese-Chat-v2, we obtained corresponding domain-specific models. Based on the initials of Motivational Interviewing and Large Language Model, we named them MI-LLMs.

**Automated and manual evaluation**

We used the methods of both automatic evaluation and manual evaluation to test the ability to use MI in conversational interviews after fine-tuning the LLMs.

**Automatic evaluation:** Because MI is a complete and progressive psychological counseling method, evaluating whether an MI-LLM adopts the MI approach requires comparing its outputs with the counselor's remarks in each round of the conversation. Therefore, we adopted the method of dividing rounds to construct the test samples, in order to obtain the output of LLM in each round of conversation.

Our test dataset consisted of complete, transcribed MI-style psychological counseling dialogues, with each dialogue corresponding to one full client–counselor session. We refer to these full dialogues as test dialogues. When constructing the round-based test samples, we split each transcribed MI-style dialogue into rounds, taking each client utterance as the model input and using all preceding turns as dialogue history to provide context. One round was defined as a single client–counselor exchange. For every round-level test sample, we attached a fixed instruction prompt: "You are a psychological counselor with 20 years of experience. Your aim is to help visitors solve psychological problems through professional Motivational Interviewing counseling."

Formally, for a test dialogue with i round (i.e., i client–counselor pairs), we denote it as $\{(q_k, r_k)|k = 1,2, ..., i\}$, where $q_k$ and $r_k$ are the client's utterance and the counselor's reply at round $k$, respectively. At round $k$, the dialogue history is defined as

$$h_k = \{(q_j, r_j)|j = 1,2, ..., k-1\}$$

$h_k$ representing all previous client queries and counselor responses. Conditioning on the

fixed MI prompt $P_{MI}$, the model generates a response $\hat{r}_k$ according to

$$\hat{r}_k = \begin{cases} f_{LLM}(P_{MI}, q_k), & k = 1, \\ f_{LLM}(P_{MI}, h_k, q_k), & 1 < k \leq i, \end{cases}$$

where $f_{LLM}(\cdot)$ denotes the inference process of the language model. Thus, each test dialogue in the test dataset gives rise to i round-level test samples in an Alpaca-style triplet form(instruction, output and input including the history and the most recent utterance from the client).

For example, in a three-round smoking-cessation conversation, Round 1 uses the client's first utterance ("Hello, I feel that I have been under a lot of pressure recently and I can't help smoking.") as the input together with the fixed MI prompt, and the corresponding counselor reply as the target output. In Round 2, the input consists of the client's second utterance ("Yes, I really want to quit smoking… but I just can't quit it."), the same fixed prompt, and the full dialogue history from Round 1 (the client's first utterance and the counselor's response), with the counselor's second reply as the target. In Round 3, the client's third utterance expressing worry about withdrawal reactions is combined with the same fixed prompt and the cumulative history from Rounds 1 and 2, and the counselor's third reply serves as the target output. A detailed example of this round-based division is shown in Supplemental Table 2.

We first fed the round-based test samples constructed above into each model to obtain its generated responses $\hat{r}_k$ for every turn $k$. For automatic evaluation, we then compared each generated response $\hat{r}_k$ with its corresponding reference counselor utterance $r_k$. The outputs of the original base models and the MI-LLMs were treated as generated texts, while the counselor's utterances in the test dataset served as reference texts. Using LLaMA-Factory, we computed BLEU-4, ROUGE-1, ROUGE-2, and ROUGE-L between each generated response and its corresponding reference. These automatic metrics were used to quantify the improvement in the models' ability to produce MI-consistent counseling responses that more closely match real counselors' language and better handle multi-turn psychological dialogues. Detailed definitions and implementation settings of the automatic metrics are provided in Supplemental Note 3.

**Manual evaluation:** In the manual evaluation, the researchers played the role of people with undesirable behaviors such as insomnia, weight loss, and addiction to mobile phones, and were randomly assigned different baseline motivation levels. For each MI-LLM, we obtained 10 independent and complete multi-round conversations, resulting in a total of 30 simulated counseling dialogues across the three MI-LLMs.

In addition, we took real-life MI psychological counseling as a reference from the AnnoMI dataset—a publicly available collection of expert-annotated counseling dialogues[38]. From AnnoMI, we randomly sampled a total of 30 motivational interviewing dialogues with real clinical records. These dialogues were evaluated manually and served as positive controls to compare with simulated counseling dialogues, in order to assess the extent to which MI-LLMs have mastered motivational interviewing.

A total of 60 multi-round dialogues, including 30 from three MI-LLMs and 30 real MI dialogues, were evaluated manually. We categorized and counted the topics in 30 simulated counseling dialogues and 30 real motivational interviewing dialogues.

We referred to MITI Coding Manual 4.2.1[29] to conduct blind manual evaluations. All MITI coding was carried out by two graduate students, one in psychology and one in medicine, both working independently under the supervision of an experienced MI clinician. The evaluation indicators included (1) global scores on the four dimensions of cultivating change talk, softening sustain talk, partnership, and empathy; (2)behavior counts including giving information, persuading with permission, asking questions, simple reflections, complex reflections, affirming, seeking collaboration, emphasizing autonomy, persuading, and confronting. The detailed explanations of overall scores and behaviors can be found in the Supplemental Table 3 and Supplemental Table 4. After completing these indicators assessments above, we used the following equation to calculate summary scores for quantitative evaluation. The assignment of points for each indicator was shown in Table 2.

For the real MI dialogues and for the dialogues generated by each MI-LLM, we separately calculated the mean and interquartile range (IQR) of the global scores, behavior counts, and summary scores. These descriptive statistics allowed us to directly compare the counseling performance of human MI practitioners with that of the three MI-LLMs.

Table 2 Method for calculating summary scores in manual evaluation

| Indicator | Calculating method |
| --- | --- |
| Complex Reflections Ratio | =Complex reflection /(Simple reflections+ Complex reflection) |
| Reflection-to-Question Ratio(R:Q) | =Total reflections/ (Total Questions)=(Simple reflections+ Complex reflection)/ (Total Questions) |
| Technical Global | =(Cultivating Change Talk + Softening Sustain Talk) / 2 |
| Relational Global | =(Partnership + Empathy) / 2 |
| Total MI-Adherent Ratio[a] | =(Seeking Collaboration + Affirming + Emphasizing Autonomy)/(Seeking Collaboration + Affirming + Emphasizing Autonomy+ Confronting + Persuading) |

[a] MITI Coding Manual 4.2.1 employed two indicators to evaluate the conformity and non-conformity of MI.

Total MI-Adherent = Seeking Collaboration + Affirm + Emphasizing Autonomy.

Total MI Non-Adherent = Confront + Persuade

However, these two indicators are related to the length of the dialogue. The longer the dialogue, the larger the values of these two indicators might be. Therefore, we adopted a ratio to normalize and compare the conformity of MI for different dialogue lengths.

Total MI-Adherent Ratio=Total MI-Adherent /(Total MI-Adherent +Total MI Non-Adherent

## Results

### Results of screening Chinese Psychological Counseling Dialogue Datasets

We obtained five Chinese psychological counseling dialogue datasets: CPsyCounD[30], Emotional First Aid Dataset[39], Psy-Insight[40], PsyDTCorpus[41], and Smilechat[42]. The quality of these datasets was evaluated from four perspectives: comprehensiveness, professionalism, authenticity, and safety. Results are shown in Figure 2. Among the five datasets, CPsyCounD and PsyDTCorpus achieved the two highest overall scores. Therefore, we constructed the training and test datasets based on these two datasets.

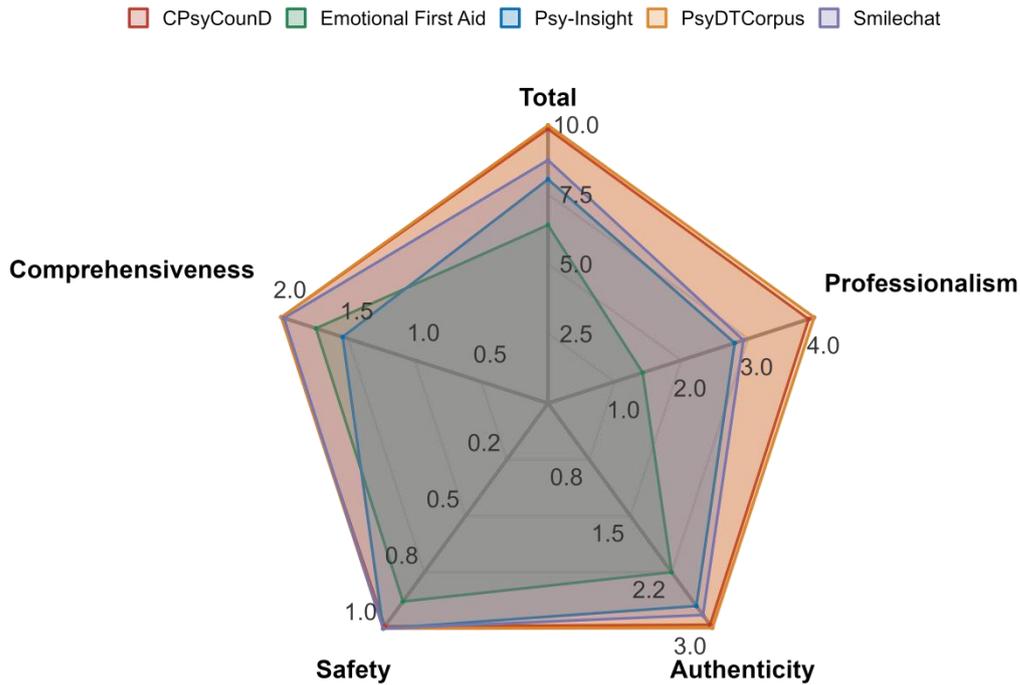

**Figure 2** The evaluation results of GPT-4 on the scores of four perspectives and the total score of five Chinese psychological counseling dialogue datasets

**Construction of MI-style Datasets**

First, we extracted 1520 multi-round dialogues from CPsyCounD and 520 dialogues from PsyDTCorpus, based on their optimal quality compared with the other three datasets. Second, we transcribed them into MI-style dialogues by using prompt engineering with GPT-4 obtaining 2,040 transcribed multi-turn dialogues in total. Third, from 2040 transcribed multi-turn dialogues, we randomly selected 2,000 of them as the training dataset and the remaining 40 as the test dataset.

Specifically, for CPsyCounD, PsyDTCorpus, the training dataset and the test dataset, we calculated the average number of dialogue rounds, the average number of words in each client utterance, and the average number of words in each counselor utterance. The results are shown in Table 3.

**Table 3** Statistical analysis of dialogue features in CPsyCounD, PsyDTCorpus, and the

MI-style training dataset

| Datasets | Average number of dialogue rounds | Average number of words in client's each utterance | Average number of words in counselor's each utterance |
| --- | --- | --- | --- |
| CPsyCounD | 7.65 | 29.95 | 47.76 |
| PsyDTCorpus | 18.0 | 28.67 | 53.74 |
| Training dataset | 7.97 | 29.08 | 57.33 |
| Test dataset | 7.53 | 29.21 | 56.42 |

**Results of Automatic Evaluation**

We divided the 40 test dialogues into 367 round-based samples to evaluate the fine-tuning effect of the model. Based on these round-based samples, automatic metrics were calculated to quantify the improvement in the models' ability to produce MI-consistent counseling responses that more closely match real counselors' language and better handle multi-turn psychological dialogues.

As shown in Figure 3, the results of fine-tuned MI-LLMs performed substantially better than those produced by the original models. Specifically, the BLEU-4 score of Baichuan2-7B-Chat–based MI-LLM increased from 2.10 for the original Baichuan2-7B-Chat to 5.84; for ChatGLM-4-9B-Chat–based MI-LLM, it increased from 1.77 to 6.47; and for Llama-3-8B-Chinese-Chat-v2–based MI-LLM it increased from 2.39 to 5.79. Regarding the ROUGE series metrics (ROUGE-1, ROUGE-2, ROUGE-L), the performance of the MI-LLMs was likewise consistently better than that of the original models.

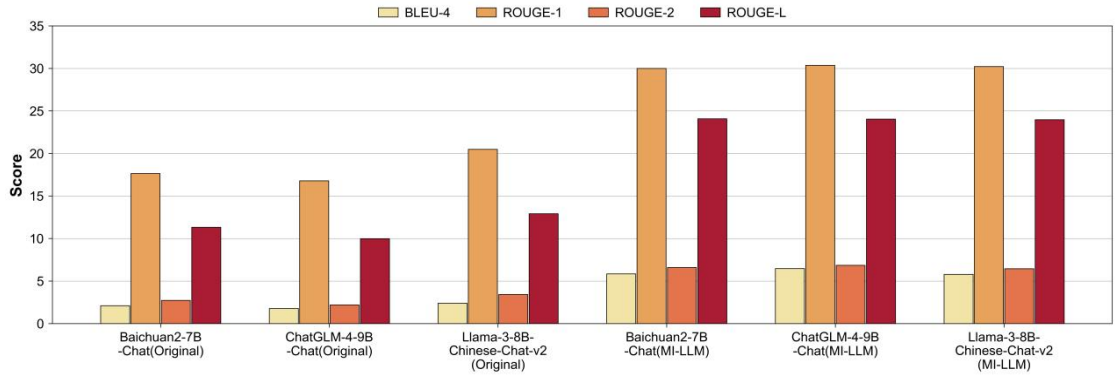

**Figure 3** The automatic evaluation results of MI-LLMs and the original model on the test datasets

### Results of Manual Evaluation

We compared the number of dialogue rounds and the average utterance length of clients and counselors in real versus simulated dialogues. As shown in Table 4, real MI dialogues contained substantially more rounds, but the counselor's utterances were noticeably shorter on average than those in the simulated dialogues. This pattern suggests that human counselors tend to intervene with briefer, more frequent turns, whereas MI-LLMs generate dialogues with fewer turns but longer counselor responses in each round.

**Table 4** Statistical analysis of dialogue features in real MI dialogues and Simulated counseling dialogues

| Source | Number of Dialogues | Average number of dialogue rounds | Average number of words in client's each utterance | Average number of words in counselor's each utterance |
|---|---|---|---|---|
| **Real MI dialogues** | 30 | 57.93 | 24.26 | 21.38 |
| **Simulated counseling dialogues from three MI-LLMs** | | | | |
| Baichuan2-7B-Chat-based | 10 | 18.3 | 49.11 | 14.8 |
| ChatGLM-4-9B-Chat-based | 10 | 17.4 | 52.28 | 11.45 |
| Llama-3-8B-Chinese-Chat-v2-based | 10 | 18.9 | 43.79 | 10.71 |

We categorized the topics of 30 real MI dialogues and 30 simulated counseling dialogues from three MI-LLMs. Both the real MI dialogues and the simulated counseling dialogues generated by the MI-LLMs primarily focused on behavior change. In particular, topics such as weight loss/diet management, increasing exercise, and reducing dependence on external objects or substances (e.g., mobile phones, alcohol, or other addictive behaviors) were represented in both sets of dialogues. Details are shown in Table 5.

**Table 5** Topics of real MI dialogues and simulated counseling dialogues from MI-LLMs

| Topics | Counts |
| --- | --- |
| **Real MI dialogues** | |
| Increase exercise | 6 |
| Chronic disease self-management | 6 |
| Weight loss/diet management | 3 |
| Reduce recidivism | 3 |
| Reduce drug use | 2 |
| Reduce alcohol use | 2 |
| Quit smoking | 1 |
| Other | 7 |
| **Simulated counseling dialogues from three MI-LLMs** | |
| Weight loss/diet management | 4×3 |
| Reducing mobile phone use | 2×3 |
| Improving insomnia | 2×3 |
| Increase exercise | 1×3 |
| Controlling advance consumption | 1×3 |

Based on the MITI framework, the manual evaluation results for simulated counseling dialogues generated by MI-LLMs and for real MI dialogues—including the mean and interquartile range for each indicator—are summarized in Table 6.

In terms of global scores, the simulated dialogues all received consistently high global ratings on cultivating change talk, softening sustain talk, empathy, and partnership, with mean scores generally above 3.5 and relatively narrow IQRs. Among them, the MI-LLM based on ChatGLM-4-9B-Chat achieved the strongest overall global performance. It reached a mean of 4.00 on all four global items and obtained Technical Global and Relational Global scores of 4.00, slightly surpassing the other two MI-LLMs and the real MI dialogues. Real MI dialogues achieved empathy scores that were marginally higher than those of the MI-LLMs, indicating that experienced human counselors still hold an advantage in conveying empathic understanding. For cultivating change talk and partnership, however, the average scores of human counselors were similar to those of MI-LLMs, suggesting that the models can already approximate human-level performance on key MI-consistent global dimensions.

In terms of behavior counts, asking questions and simple reflections were the two most frequent behaviors in both MI-LLM–generated and real MI dialogues. Complex reflections occurred most often in real MI dialogues, with Baichuan2-7B-Chat–based and Llama-3-8B-Chinese-Chat-v2–based MI-LLMs producing somewhat more complex reflections than the ChatGLM-4-9B-Chat–based model, but still slightly fewer than human counselors on average.

Compared with real MI dialogues, MI-LLMs produced substantially more instances of giving information and affirming, indicating a strong tendency to support clients through explanations and positive feedback. Certain MI-adherent behaviors such as seeking collaboration and emphasizing autonomy appeared at least as frequently in MI-LLM simulations as in real dialogues, whereas persuading with permission was more common in real MI dialogues. MI non-adherent behaviors, including persuading and confronting, were infrequent in both groups, and confronting was almost absent.

The dispersion of behavior counts also differed between groups: dialogues generated by MI-LLMs showed relatively stable behavior frequencies across conversations (narrower IQRs), whereas real MI dialogues exhibited greater variability between dialogues, reflecting the flexible adaptation of strategies in real clinical practice.

In terms of summary scores, real MI dialogues showed a clear advantage in complex reflective skills. The complex reflections ratio was highest in real MI dialogues, exceeding the ratios observed in all three MI-LLMs, which suggests that human counselors are more likely to use reflections that add inference, meaning, or emotional elaboration rather than merely repeating content. Similarly, the reflection-to-question (R:Q) ratio was higher in real MI dialogues than in any of the MI-LLM–generated dialogues, indicating that human practitioners rely relatively more on reflective listening than on asking questions.

In contrast, MI-LLMs performed very well on MI-adherence. The total MI-adherent ratio was highest for the ChatGLM-4-9B-Chat–based MI-LLM, followed by the Baichuan2-7B-Chat–based model, both at or above the level observed in real MI dialogues, while the Llama-3-8B-Chinese-Chat-v2–based model was slightly lower but still close. Taken together, these results suggest that MI-LLMs are highly capable of following MI-consistent guidelines and avoiding non-adherent behaviors, whereas human counselors retain an advantage in more nuanced skills, particularly the use of complex reflections and maintaining a higher balance of reflections over questions.

**Table 6** The manual evaluation results of global scores, behavior counts and summary scores

| Evaluation indicators | Simulated counseling dialogues from MI-LLMs | | | Real MI dialogues |
|---|---|---|---|---|
| | Baichuan2-7B-Chat-based | ChatGLM-4-9B-Chat-based | Llama-3-8B-Chinese-Chat-v2-based | |
| **Global scores** | | | | |
| Cultivating change talk | 4.00(4.00,4.00) | 4.00(4.00,4.00) | 4.00(4.00,4.00) | 3.97(3.25, 5.00) |
| Softening sustain talk | 3.89(4.00,4.00) | 4.00(4.00,4.00) | 3.90(4.00,4.00) | 3.87(3.00, 4.00) |
| Empathy | 3.78(4.00,4.00) | 4.00(4.00,4.00) | 3.80(4.00,4.00) | 4.07(4.00, 5.00) |
| Partnership | 3.44(3.00,4.00) | 4.00(4.00,4.00) | 4.00(4.00,4.00) | 3.83(3.00, 4.00) |
| **Behavior counts** | | | | |
| Giving information | 3.22(2.00,4.00) | 4.00(3.00,5.75) | 4.50(3.25,5.75) | 1.07(0.00, 1.75) |

| Evaluation indicators | Simulated counseling dialogues from MI-LLMs | | | Real MI dialogues |
|---|---|---|---|---|
| | Baichuan2-7B-Chat-based | ChatGLM-4-9B-Chat-based | Llama-3-8B-Chinese-Chat-v2-based | |
| Persuading with permission | 0.33(0.00,1.00) | 0.00(0.00,0.00) | 0.10(0.00,0.00) | 0.80(0.00, 0.00) |
| Asking questions | 15.56(9.50,22.00) | 14.20(11.75,15.38) | 16.00(12.75,17.75) | 13.43(5.25,18.25) |
| Simple reflections | 13.50(8.50,19.00) | 13.25(12.12,14.25) | 13.10(11.00,14.88) | 14.17(1.25,18.50) |
| Complex reflections | 6.00(4.00,8.00) | 4.25(2.50,5.75) | 5.00(3.50,5.62) | 6.13(1.00, 5.75) |
| Total reflections | 19.5(12.50,27.00) | 17.5(15.50,18.50) | 18.1(14.25,21.75) | 20.3(6.25, 22.00) |
| Affirming | 13.61(10.50,18.50) | 11.45(8.88,12.38) | 13.75(10.12,14.25) | 3.30(0.25, 3.75) |
| Seeking collaboration | 3.50(2.50,4.00) | 2.80(2.50,3.00) | 2.75(2.00,3.50) | 1.53(0.25, 2.00) |
| Emphasizing autonomy | 1.67(1.50,2.00) | 1.60(1.12,2.00) | 0.90(0.12,1.38) | 0.60(0.00, 1.00) |
| Persuading | 3.83(2.50,4.00) | 2.30(1.50,3.38) | 3.40(2.50,4.38) | 1.57(0.00, 3.00) |
| Confronting | 0.28(0.00,0.50) | 0.25(0.00,0.50) | 0.15(0.00,0.38) | 0.10(0.00, 0.00) |
| **Summary scores** | | | | |
| Complex Reflections Ratio | 0.31(0.29,0.32) | 0.24(0.22,0.29) | 0.26(0.23,0.30) | **0.37(0.06, 0.72)** |
| Reflection-to-Question Ratio (R:Q) | 1.27(1.21,1.33) | 1.25(1.20,1.26) | 1.11(1.01,1.21) | **1.44(0.83, 1.90)** |
| Technical Global | 3.94(4.00,4.00) | **4.00(4.00,4.00)** | 3.95(4.00,4.00) | 3.92(3.63, 4.50) |
| Relational Global | 3.61(3.50,4.00) | **4.00(4.00,4.00)** | 3.90(3.62,4.00) | 3.95(3.50, 4.50) |
| Total MI-Adherent | 0.75(0.71,0.79) | **0.78(0.67,0.83)** | 0.72(0.69,0.75) | 0.73(0.62, 1.00) |

| Evaluation indicators | Simulated counseling dialogues from MI-LLMs | | | Real MI dialogues |
|---|---|---|---|---|
| | Baichuan2-7B-Chat-based | ChatGLM-4-9B-Chat-based | Llama-3-8B-Chinese-Chat-v2-based | |
| Ratio | | | | |

**Discussion**

MI-consistent counseling skills are well recognized as important in multiple fields of health promotion, but mastering MI skills is time-consuming and infeasible for most health practitioners. This exploratory study tested the feasibility of training LLMs to approximate core MI-consistent skills demonstrated by human experts. Both the automatic and manual evaluations suggested that MI-LLMs had largely acquired core MI-consistent techniques and principles. They achieved global technical and relational scores and MI-adherent ratios that were close to those of real counselors, although human practitioners still showed clear advantages in complex reflections and in maintaining a higher reflection-to-question ratio. Taken together, these findings represent an initial but informative step toward integrating LLMs with MI-based counseling.

In recent years, the application of LLMs in the field of behavior change has developed rapidly, giving rise to various types of conversational agents or chatbots[41, 43, 44], including emerging systems that explicitly take MI as their core framework. For example, health-coaching systems built on GPT-series models (such as GPTCoach)[45], as well as other LLM-based health coaching studies, have shown that LLMs can generate empathetic and contextually appropriate counseling responses, which can approach or even match human coaches. In specific behavioral domains, Brown and colleagues developed an MI-based chatbot for smoking cessation, which enhanced smokers' readiness to quit by generating reflective listening responses, suggesting that in specific clinical contexts LLMs can be steered to produce MI-consistent reflections and responses[46]. Reviews on LLM-based health coaching and behavior change applications have also noted that, although this field is expanding rapidly, most interventions remain exploratory, with considerable heterogeneity in design and evaluation methods[47, 48].

Building on this body of work, the present study extends the literature in several ways and exhibits several strengths. First, to our knowledge, it is among the earliest attempts to systematically develop and evaluate a MI–oriented LLM in Chinese, addressing a major gap in the literature that is currently dominated by English-language systems. By leveraging GPT-4 to transcribe ordinary psychological counseling dialogues into MI-style conversations, we propose a scalable and practically implementable pathway for constructing MI-style training data in contexts where large expert-annotated MI corpora are unavailable or infeasible to obtain. Second, we employed a dual evaluation strategy that combines automatic metrics with a rigorous manual assessment grounded in the MITI framework. This enables us to move beyond global impressions or user satisfaction and examine specific counselor behaviors—such as reflection-to-question ratio, complex reflection ratio, and overall MI adherence—providing a fine-grained understanding of what MI-LLMs have learned and where gaps remain. Third, by directly comparing MI-LLMs with real counselors on key MITI indicators, rather than only contrasting different models with each other, we offer a more clinically meaningful benchmark for interpreting model performance. Finally, the study integrates methodological innovation (prompt-based MI transcription, targeted fine-tuning, and MITI-based evaluation) with a clear application focus on health-related risk behaviors, thereby demonstrating both the technical feasibility and the applied relevance of MI-LLMs as a potential tool for health promotion and behavior change.

Despite the completion of the dataset construction, model fine-tuning, and evaluation, several limitations remained in the present study. First, although the training dataset was of high quality, it was relatively small (only 2,000 multi-turn dialogues), which may lead to limited generalization ability and underfitting. Manual evaluation revealed that even the best-performing ChatGLM-4-9B-Chat–based MI-LLM struggled to fully master complex MI techniques such as complex reflection, often producing responses limited to paraphrasing or slight reformulation, failing to deeply capture clients' implicit emotions or internal conflicts. Second, while both automatic and manual evaluations indicate that MI-LLMs acquired basic MI skills and could appropriately apply key techniques in risk-related dialogues, our study did not include randomized controlled trials (RCTs) or other rigorous intervention designs. As a result, we cannot draw definitive conclusions about their real-world effectiveness for behavior change; rather, the current findings should be viewed as providing preliminary evidence of feasibility and potential. Finally, the "black-box" nature of LLMs remains an inherent challenge. Although MI-LLMs performed well on multiple indicators, the internal reasoning behind their responses is not fully transparent. This issue is shared by most contemporary LLM-based mental health applications and highlights the broader need for more interpretable and controllable models.

To address these limitations, future work could further proceed in several directions. First, expand the scale and diversity of datasets: Enhance training and generalization by collecting real counseling dialogues, applying data augmentation techniques such as synonym replacement and syntactic transformation, and transcribing existing counseling records to enrich MI datasets. Second, introduce reinforcement learning from human feedback (RLHF)[49, 50]: RLHF allows models to optimize outputs based on human preferences. Future studies could, under the guidance of MI experts, collect fine-grained rankings or ratings of multiple candidate responses generated by MI-LLMs, focusing on key MI techniques such as emotional reflection, open-ended questions, and resolving ambivalence. These preference datasets can be used to train a reward model and optimize the generation policy through the RLHF framework, enabling MI-LLMs to better master complex interactive MI skills and produce responses that meet professional standards. Third, design rigorous RCTs: RCTs should be conducted to evaluate MI-LLMs' effectiveness in real-world interventions and assess their long-term impact on behavior change. This would provide stronger evidence for their practical utility. Fourth, enhance transparency and interpretability: Future research could incorporate explicit reasoning frameworks, in which responses are generated through intermediate, interpretable reasoning steps before producing the final answer. This involves decomposing the model's decision-making into a series of interpretable, logically clear intermediate steps. For example, the process of evoking behavioral change could be broken down into motivation identification, ambivalence analysis, and response generation. Making the model's reasoning visible enhances understanding, trust, and acceptance among clients and counselors.

In conclusion, our initial exploration of MI-LLMs not only suggests that MI-LLMs emulate the MI capabilities of human counselors on MITI-based process measures but also shows potential in evoking motivation for change and promoting healthy behavior. Findings from this study could pave the way for further research in terms of data scale, training methods, and real-world applications. Future efforts should focus on improving MI-LLM's generalization, transparency, and practicality to facilitate their broader adoption in the health domain.